%% file: main.tex
\documentclass{article}
\usepackage{iclr2027_conference,times}
\input{math_commands.tex}

\usepackage{amsmath,amssymb}
\usepackage{booktabs}
\usepackage{graphicx}
\usepackage{float}
\usepackage{wrapfig}
\usepackage{rotating}
\usepackage{xcolor}
\usepackage{hyperref}
\usepackage{url}
\AddToHook{env/table/begin}{%
  \setlength{\abovecaptionskip}{6pt}%
  \setlength{\belowcaptionskip}{2pt}%
  \vspace{2pt}}
\AddToHook{env/table/end}{\vspace{2pt}}
\hypersetup{
  hidelinks,
  pdftitle={VASC: Value-Aware Sparse Attention with Cross-Layer Memory for Efficient 3D Reconstruction},
  pdfauthor={Junyi Wu, Fanqing Kong, Leyang Chen, Shaoqiu Zhang, Yulun Zhang},
  pdfsubject={Training-free sparse attention for efficient 3D reconstruction},
  pdfkeywords={sparse attention, 3D reconstruction, cross-layer memory}
}

\title{\raggedright VASC: Value-Aware Sparse Attention\\
with Cross-Layer Memory\\
for Efficient 3D Reconstruction}

\author{%
Junyi Wu\thanks{Equal contribution.},\enspace
Fanqing Kong\footnotemark[1],\enspace
Leyang Chen\footnotemark[1],\enspace
Shaoqiu Zhang,\enspace
Yulun Zhang\thanks{Corresponding author: Yulun Zhang,
\href{mailto:yulun100@gmail.com}{yulun100@gmail.com}}\\
\normalfont Shanghai Jiao Tong University}

\iclrfinalcopy
\begin{document}
\raggedbottom
\maketitle
\pagestyle{plain}
\vspace{-18pt}

\begin{figure}[H]
\centering
\includegraphics[width=\textwidth]{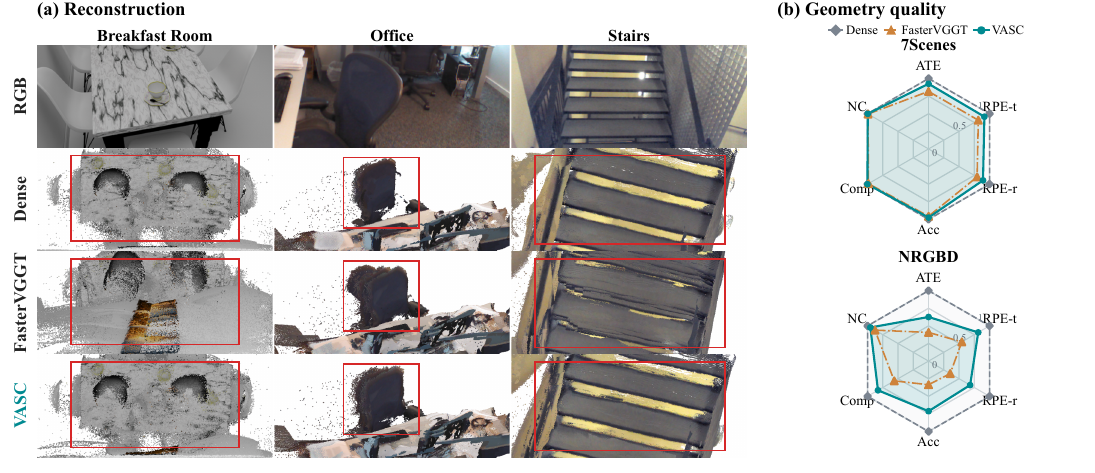}
\caption{\textbf{Reconstruction quality with sparse attention.}
Left: RGB references and aligned stride-10 reconstructions; red boxes mark
shared regions. For each scene, FasterVGGT and VASC use the same sparsity.
RGB and rendered views are not pixel-aligned.
Right: dataset-level metrics normalized to dense VGGT. Higher values indicate
lower errors or higher normal consistency (NC).}
\label{fig:teaser}
\end{figure}

\begin{abstract}
Feed-forward 3D vision models such as VGGT have achieved remarkable progress,
unifying camera estimation and dense scene reconstruction in a single pass.
However, their quadratic global attention makes long image sequences expensive,
while existing sparse methods may favor highly attended yet value-redundant regions.
To address these limitations, we introduce \textbf{VASC}, a training-free sparse
attention method combining \textbf{value-aware block selection} and
\textbf{execution-aware cross-layer memory}. Our value-aware block selection
integrates pooled query--key relevance with neighboring value contrast, reducing
redundancy while preserving query-relevant and distinctive content.
Cross-layer memory tracks unserved demand across layers and updates this state
according to actual execution, enabling previously underserved blocks to compete
under a fixed computation budget. Experiments on 7Scenes and NeuralRGB-D with VGGT
and ${\pi}^3$ demonstrate improved pose estimation and reconstruction quality
compared with FasterVGGT, together with up to $2.29\times$ faster inference than
dense VGGT. Code is available at \url{https://github.com/kosakayamahoo-design/VASC}.
\end{abstract}

\input{sections/01_introduction}
\clearpage
\input{sections/02_related_work}
\clearpage
\input{sections/03_observation}
\clearpage
\begingroup
\setlength{\parskip}{3pt}
\input{sections/03_method}
\par
\endgroup
\input{sections/04_experiments}

\input{sections/05_conclusion}

\clearpage
\bibliography{iclr2027_conference}
\bibliographystyle{iclr2027_conference}

\appendix
\clearpage
\paragraph{Organization of the appendix.}
Appendix~\ref{app:scheduler-details} gives implementation details;
Appendix~\ref{app:observation-protocols} describes observation and rendering
protocols. Benchmark results, ablations, and backbone adaptations appear
in Appendices~\ref{app:additional-experiments},
\ref{app:ablation-studies}, and~\ref{app:host-adapters}, respectively.

\input{sections/appendix_a_implementation}
\input{sections/appendix_b_observation_protocols}
\input{sections/appendix_c_benchmarks}
\input{sections/appendix_d_ablations}

\input{sections/appendix_e_adapters}

\end{document}

%% file: math_commands.tex
\usepackage{amsmath,amsfonts,bm}

\def\eqref#1{equation~\ref{#1}}

\def\1{\bm{1}}

\DeclareMathAlphabet{\mathsfit}{\encodingdefault}{\sfdefault}{m}{sl}
\SetMathAlphabet{\mathsfit}{bold}{\encodingdefault}{\sfdefault}{bx}{n}



%% file: sections/01_introduction.tex
\section{Introduction}
\label{sec:intro}

Recovering three-dimensional geometry from images is increasingly becoming
a problem of learned, feed-forward inference. Building on transformer
representations \citep{vaswani2017attention,dosovitskiy2020image}, DUSt3R casts pairwise geometry as
pointmap regression, while MASt3R connects this representation to dense
matching \citep{wang2024dust3r,leroy2024grounding}. VGGT jointly predicts cameras, depth,
point maps, and tracks, and ${\pi}^3$ develops permutation-equivariant
geometry prediction \citep{wang2025vggt,wang2026pi}. Scaling these models
requires both accurate predictions and efficient processing of long inputs.

Global cross-view attention is a central obstacle to this scalability:
its interactions grow quadratically with the number of visual tokens.
Efficient exact-attention kernels reduce memory traffic and improve
hardware utilization \citep{dao2022flashattention,dao2024flashattention}, but do not
decide which interactions can be omitted. Recurrent and streaming models
address long inputs through spatial memory, persistent state, or causal
processing \citep{wang20253d,wang2025continuous,wu2026point3r,zhuo2026streaming}; TTT3R and InfiniteVGGT
further study long-term memory \citep{chen2026ttt3r,yuan2026infinitevggt}. We consider a
complementary setting: accelerating an existing backbone's global
attention without retraining or replacing its prediction architecture.

Training-free acceleration offers several ways to reduce this cost.
Token merging consolidates similar representations
\citep{bolya2022token,shen2025fastvggt}; sparse global attention retains
selected block interactions \citep{wang2026block}; layer scheduling changes
where cross-view computation is spent \citep{sun2026avggt}
(see also geometry quantization and video-generation caching;
\citealp{feng2026quantized,wu2025quantcache}). For sparse execution, a key
question remains: which interactions should receive a fixed computation
budget? Selection must protect geometry prediction without an overhead
that erases the savings.

Pooled query--key scores provide a natural relevance estimate, but sparse
selection faces two linked limitations in content and execution history.
Attention weights describe query-dependent relevance, whereas value
vectors supply the aggregated content, a distinction also explored in
value-aware approximation \citep{gupta2021value}. Strong attention can
coincide with locally similar values, so relevance alone does not indicate
how distinct the selected content is. Under a tight budget, relevant but
distinctive interactions may consequently be omitted, as illustrated by
the attention ranking in Figure~\ref{fig:motivation-observation}. This limitation
also extends across network depth: independent selection at each layer
does not account for whether previously assigned demand was actually
served. Historical scores and reusable sparse indices provide useful
context \citep{zhang2023h2o,bai2026indexcache,han2026evolving}, but do not
by themselves encode the gap between requested and executed computation.
The challenge is therefore to account for both value content and unmet
computational demand while preserving the efficiency of fixed-budget
sparse attention.
Sparse attention must account for both what deserves computation and
whether previously assigned demand was actually served.

We introduce \textbf{VASC}, \textbf{Value-Aware Sparse Attention with
Cross-Layer Memory}, to address these two aspects jointly. Value-aware
selection combines pooled query--key relevance with the RMS contrast of
adjacent pooled value blocks, distinguishing content at native
sparse-block granularity. Cross-layer memory maintains one scalar balance
per block pair, accumulating unserved demand and subtracting actual
execution. This computational deficit participates in subsequent ranking
alongside current demand. Both components preserve per-query capacity, the native sparse
kernel, special-token handling, and geometry heads.

Experiments on 7Scenes and NeuralRGB-D evaluate VGGT and ${\pi}^3$ across
sparsity levels and input strides. VASC improves all six pose and
reconstruction metrics over the corresponding sparse baselines at S75 in
the main comparisons, while retaining up to $2.29\times$ speedup over dense
VGGT. Figures~\ref{fig:teaser} and~\ref{fig:nc-latency} summarize quality
and efficiency. Component ablations, denser inputs, ScanNet transfer,
and scheduler adaptations further examine effectiveness and generality.

Our contributions are fourfold:
\begin{itemize}
  \item \textbf{Observation.} We illustrate a mismatch between attention
  priority and local value variation in a geometry transformer, motivating
  their complementary use.
  \item \textbf{Value-aware selection.} We combine pooled query--key
  relevance with adjacent-value contrast for training-free selection of
  distinctive, query-relevant blocks.
  \item \textbf{Cross-layer memory.} An execution-updated pair-level state
  carries unserved demand across layers to compete with current demand
  under an unchanged budget.
  \item \textbf{Effectiveness and efficiency.} Across two geometry
  backbones, VASC improves sparse reconstruction and pose estimation,
  with controlled ablations and up to $2.29\times$ faster inference than
  dense VGGT.
\end{itemize}

%% file: sections/02_related_work.tex
\section{Related Work}
\label{sec:related}

\begin{wrapfigure}{l}{0.43\textwidth}
  \centering
  \includegraphics[width=\linewidth]{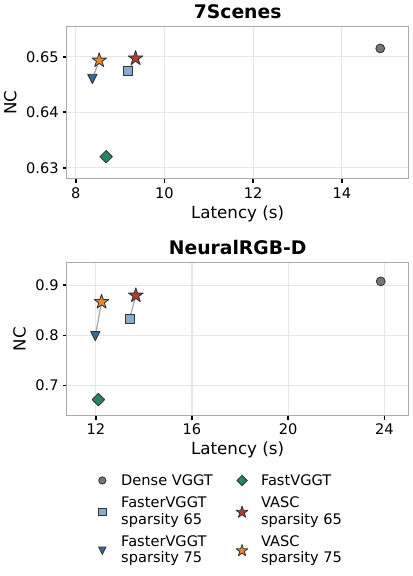}
  \caption{\textbf{Quality--latency comparison.}
  VGGT at stride 10. Higher NC and lower latency are better.}
  \label{fig:nc-latency}
\end{wrapfigure}

\paragraph{Feed-forward and streaming visual geometry.}
DUSt3R predicts pairwise pointmaps and aligns them across views;
MASt3R augments this geometry-based formulation with dense matching
\citep{wang2024dust3r,leroy2024grounding}. VGGT and ${\pi}^3$ advance joint multi-view geometry
prediction through global attention \citep{wang2025vggt,wang2026pi}.
Spann3R, CUT3R, and Point3R instead maintain spatial or recurrent state
for incremental reconstruction \citep{wang20253d,wang2025continuous,wu2026point3r}.
StreamVGGT adapts geometry prediction to streaming inputs, while TTT3R
and InfiniteVGGT address the maintenance of long-term state
\citep{zhuo2026streaming,chen2026ttt3r,yuan2026infinitevggt,wu2026balancegs}. VASC targets interaction selection inside the global
attention of pretrained, multi-view geometry backbones.

\paragraph{Efficient attention and token reduction.}
FlashAttention improves exact-attention execution through IO-aware
computation and parallelism \citep{dao2022flashattention,dao2024flashattention}.
Structured sparsity reduces the interaction set through local,
factorized, or mixed connectivity
\citep{child2019sparse,beltagy2020longformer,zaheer2020big}.
DynamicViT, EViT, and TokenLearner learn to prune, reorganize, or aggregate
visual tokens \citep{rao2021dynamicvit,liang2022not,ryoo2021tokenlearner};
ToMe merges similar representations \citep{bolya2022token}.
For language models, MInference identifies sparse attention patterns
for prefill and Quest makes sparse access query-dependent
\citep{jiang2024minference,tang2024quest}. These works motivate efficient selection but
operate at different training, token, or inference interfaces. VASC
preserves child-block identity and changes the support passed to the
existing sparse attention kernel.

\paragraph{Training-free acceleration of geometry models.}
FastVGGT adapts token merging to visual geometry, FasterVGGT uses
pooled-QK proposals and block-sparse execution, and AVGGT changes the
allocation of global-attention computation
\citep{shen2025fastvggt,wang2026block,sun2026avggt}.
QuantVGGT addresses model compression through post-training quantization
\citep{feng2026quantized}. GHOST exploits predicted geometry for online KV-token
eviction in streaming reconstruction \citep{chen2026ghost}.
VASC focuses on selecting value-sensitive
block interactions and accounting for their execution across depth.

\paragraph{Value-aware and history-aware selection.}
Value-aware Approximate Attention incorporates values when approximating
attention outputs \citep{gupta2021value}. VASC uses adjacent-value contrast
to refine pooled relevance at low cost. H$_2$O and SnapKV compress language-model
KV state using attention-derived importance, and StreamingLLM preserves
attention sinks with a recent-context window
\citep{zhang2023h2o,li2024snapkv,xiao2024efficient}. IndexCache reuses sparse indices
across layers, while EvoSparse exploits token-importance dynamics
\citep{bai2026indexcache,han2026evolving}. Our memory instead records the
gap between assigned demand and actual execution. Its nonnegative
recurrence follows classical queue accounting
\citep{lindley1952theory,neely2010stochastic}, providing an execution-aware
summary of allocation history.

%% file: sections/03_observation.tex
\section{Analyzing Attention and Value in VGGT}
\label{sec:observation}

\begin{figure}[!htbp]
  \centering
  \includegraphics[width=\textwidth,trim=24bp 35bp 24bp 0bp,clip]{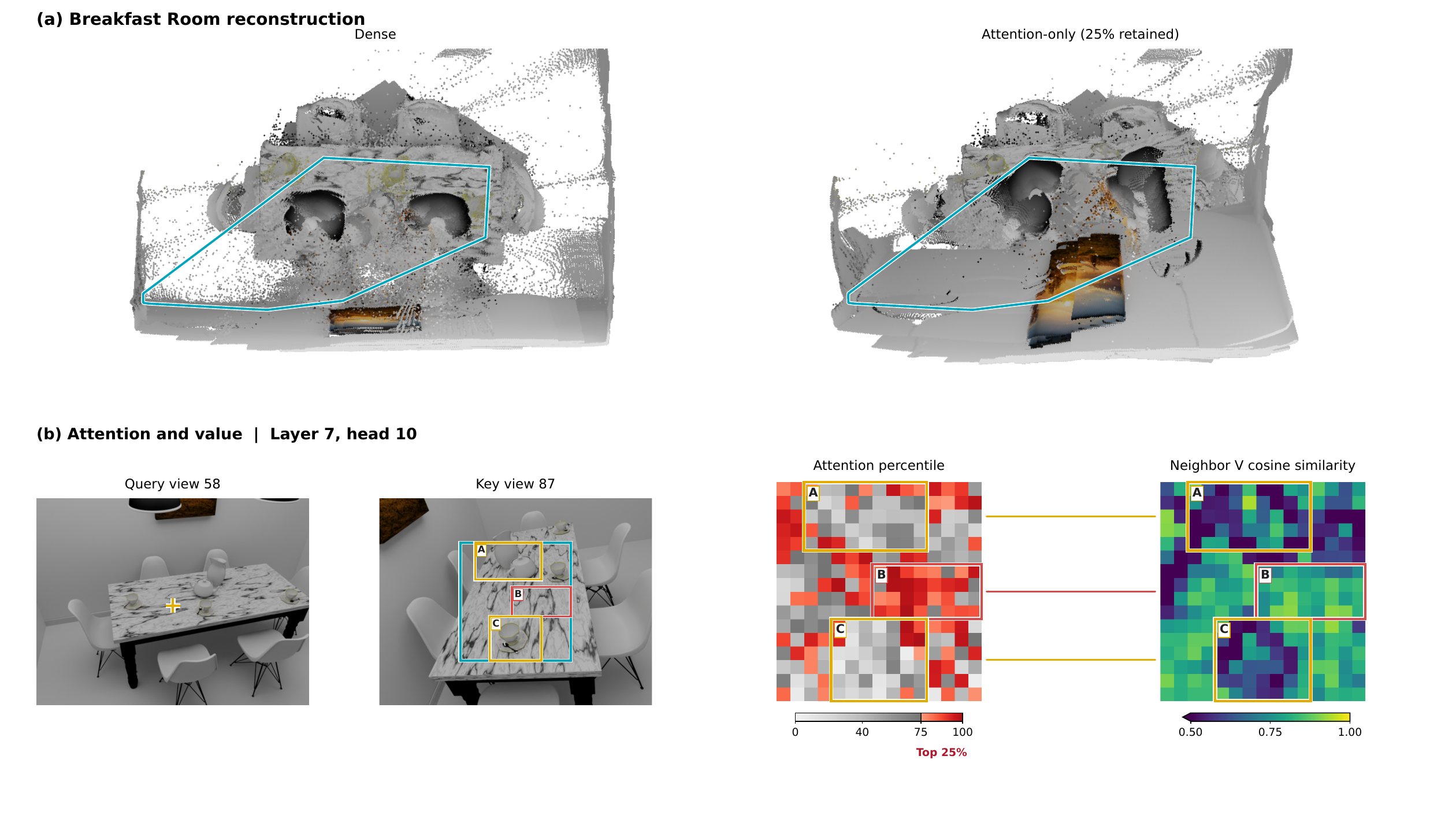}
  \caption{\textbf{Attention priority and local value variation.}
  (a) Dense and attention-only reconstruction with 25\% of connections retained.
  (b) Attention and neighboring-value similarity reveal different
  spatial patterns; lower similarity indicates greater directional variation.}
  \label{fig:motivation-observation}
\end{figure}

\paragraph{Reconstruction under attention-only selection.}
Which connections should be preserved when only a small fraction of
attention can be retained? We investigate this question through
reconstruction experiments on scenes from NeuralRGB-D and 7Scenes.
Starting from pretrained VGGT, we compare dense attention with an
attention-only setting that retains the highest-ranked 25\% of patch-key
connections for each patch query and head in the global attention layers.
The retained weights are renormalized, while frame attention, special-token
handling, and model parameters remain unchanged. As shown in
Figure~\ref{fig:motivation-observation}(a), retaining the strongest
attention responses alone can still degrade reconstruction quality.
This motivates examining both the relevance and the value content of
selected connections.

\paragraph{Attention relevance and value variation.}
We compare attention priorities with local value variation, visualized
using each value vector's mean cosine similarity to its four spatial
neighbors. Lower similarity indicates greater directional variation.
Figure~\ref{fig:motivation-observation}(b) illustrates that the two maps
need not agree: highly attended connections can carry locally similar
values, while lower-ranked connections can contain more distinct content.
Attention-only selection can therefore favor redundant content under a
tight budget. Visualization details are given in
Appendix~\ref{app:observation-protocols}.

\paragraph{Competition under a fixed budget.}
Retaining a connection consumes capacity that could otherwise serve
another candidate. A high attention rank therefore does not, by itself,
indicate how much distinct content is preserved by that choice.
Conversely, falling below the selection threshold does not imply
irrelevance to the query. The resulting trade-off concerns not only how
much attention is retained, but also which value content is represented
within the available capacity.

\paragraph{Implications for sparse selection.}
In an attention output $o_i=\sum_j a_{ij}v_j$, the weights describe
query--key relevance, while the values supply the aggregated content.
Ranking solely by $a_{ij}$ does not account for similarity among values.
This motivates \textbf{joint QKV-informed ranking}: query--key relevance
identifies relevant interactions, and value variation helps distinguish
their content. Such a ranking expresses which interactions deserve
computation; the fixed budget still determines which can actually be
executed. This distinction between demand and execution underlies VASC's
selection and feedback mechanism.

%% file: sections/03_method.tex
\begin{figure}[!t]
  \centering
  \includegraphics[width=\textwidth]{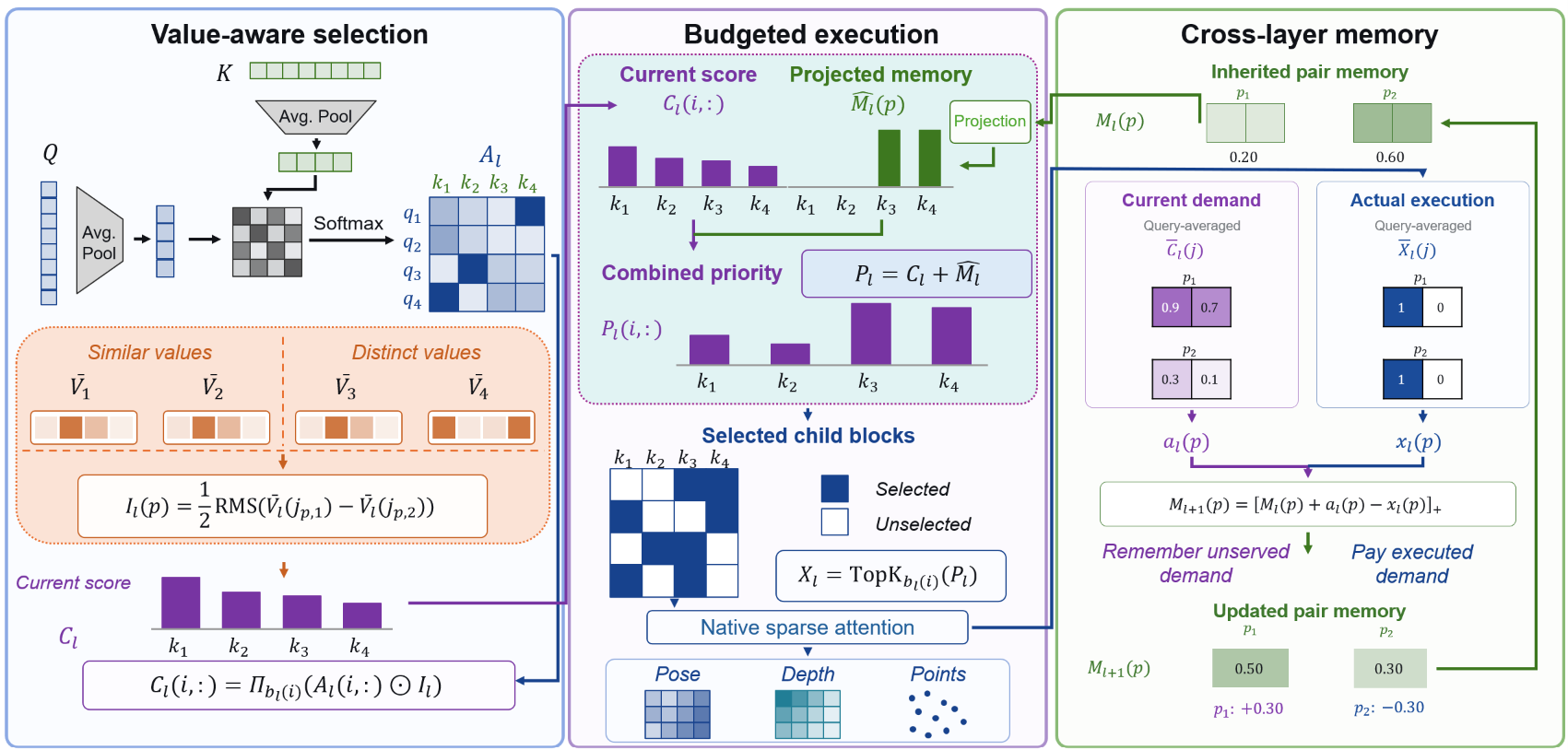}
  \caption{\textbf{VASC overview.} Value contrast and cross-layer memory
  jointly guide block selection within a fixed budget. Actual execution
  updates memory for the next layer.}
  \label{fig:method}
\end{figure}

\section{Method}
\label{sec:method}

VASC combines query--key relevance and value contrast into current demand,
allocates a fixed execution budget, and carries their accumulated mismatch
across layers (Figure~\ref{fig:method}). Its low-cost, training-free
realization preserves the native sparse attention kernel and geometry heads.

\paragraph{Blocks and capacity.}
At layer $l$, $i$ indexes a query block and $j$ a key/value child block,
the unit selected for sparse execution. Each query has capacity $b_l(i)$.
Adjacent key/value children form pairs $\mathcal C(p)$ with fixed
membership across layers; $p(j)$ denotes the pair containing child $j$.
Pairs share a value-contrast score and a memory state, but their children
remain separate execution candidates. Batch and head indices are omitted.

\subsection{Value-Aware Block Selection}

Section~\ref{sec:observation} shows that attention-only selection can
degrade reconstruction. Attention priority also differs from local value
variation: highly attended regions can contain similar values, while
lower-ranked regions carry more distinct content. This motivates combining
query relevance with value contrast.

We express this joint QKV-informed scoring as
\begin{equation}
 C_l=\mathcal F(Q_l,K_l,V_l;b_l).
 \label{eq:demand-framework}
\end{equation}
Here $C_l(i,j)$ is \textbf{normalized computation demand} for interaction
$(i,j)$. QK measures relevance to the query; value contrast distinguishes
the content from nearby alternatives.

\textbf{A low-cost realization.} VASC instantiates $\mathcal F$ with
average-pooled block features. Pooled queries and keys provide relevance:
\begin{equation}
 A_l(i,j)=\operatorname{Softmax}_{j}\!\left(
 \frac{\bar Q_l(i)^\top\bar K_l(j)}{\sqrt d}\right).
 \label{eq:qk}
\end{equation}
\begin{samepage}
For each adjacent pair, RMS contrast between pooled values supplies a
shared content signal. We use it to modulate the relevance of both children:
\begin{equation}
 \begin{aligned}
 I_l(p)&=\tfrac12\operatorname{RMS}\!\left(
 \bar V_l(j_{p,1})-\bar V_l(j_{p,2})\right),\\
 C_l(i,:)&=\Pi_{b_l(i)}\!\left(A_l(i,:)\odot I_l(p(:))\right).
 \end{aligned}
 \label{eq:pv}
\end{equation}
\end{samepage}
Budget rescaling and clipping by $\Pi_b$ give $0\leq C_l(i,j)\leq1$ and
$\sum_j C_l(i,j)\leq b_l(i)$, making demand comparable to unit execution.
Contrast is shared across queries; relevance remains query-specific.
RMS captures these value differences with a low-cost blockwise reduction.

\subsection{Budgeted Execution}

\textbf{From demand to execution.} Value-aware scores describe which
interactions deserve computation; the fixed budget determines which are
executed. To account for both current demand and past service, we combine
$C_l$ with a compact memory $M_l$ of accumulated demand--execution imbalance:
\begin{equation}
 X_l=\mathcal G(C_l,M_l;b_l),
 \qquad \sum_j X_l(i,j)=b_l(i).
 \label{eq:execution-framework}
\end{equation}
Each $X_l(i,j)\in\{0,1\}$ supplies one unit of execution when selected.
Memory redistributes execution within this fixed capacity.

\textbf{A budget-preserving realization.} VASC adds a bounded
copy $\widetilde M_l$ of pair memory to current demand, then selects
children within each query's capacity:
\begin{equation}
 \begin{aligned}
 P_l(i,j)&=C_l(i,j)+\widetilde M_l(p(j)),\\
 X_l(i,:)&=\operatorname{TopKMask}_{b_l(i)}\!\left(P_l(i,:)\right).
 \end{aligned}
 \label{eq:budget-realization}
\end{equation}
Each query selects its own children from priority $P_l$ and passes the
binary mask $X_l$ directly to the native sparse attention kernel.
We preserve the host's special-token handling: camera and register queries
use dense attention, and special-only key blocks remain always kept.
These always-kept blocks do not compete for the patch-child budget above.

\subsection{Cross-Layer Memory}

\textbf{Refine allocation through past service.} Current demand evaluates
present content; memory adjusts allocation using accumulated
demand--execution imbalance at the block-pair level. This gives previously
underserved blocks another opportunity to compete for the fixed budget.
In particular, near the selection cutoff, past service can distinguish
candidates with similar current demands.

\textbf{A compact update.} We apply the same averaging over query blocks
and each pair's children to $C_l$ and $X_l$, obtaining pair demand $a_l(p)$
and service $x_l(p)$, both in $[0,1]$. Starting from zero, we retain
their accumulated nonnegative imbalance:
\begin{equation}
 M_{l+1}(p)=[M_l(p)+a_l(p)-x_l(p)]_+,\qquad M_1(p)=0.
 \label{eq:memory}
\end{equation}
Demand adds to the balance; service subtracts from it and can reduce
earlier deficits. Clipping prevents negative memory, but receiving some
computation need not clear the balance. The resulting pair-level state
summarizes demand and service across queries and paired children.

One scalar per pair, batch entry, and head is retained for updating;
its bounded copy informs selection. Each layer uses its own Q, K, and V
to compute attention on the selected support.

%% file: sections/04_experiments.tex
\section{Experiments}
\label{sec:experiments}

\subsection{Experimental Setup}

\paragraph{Datasets and metrics.}
We evaluate VGGT and ${\pi}^3$ on all 18 test sequences of 7Scenes
\citep{shotton2013scene} and all nine NeuralRGB-D (NRGBD) sequences
\citep{azinovic2022neural} at strides $s\in\{5,10\}$, retaining every
$s$-th frame without a frame cap. Generalization uses 50 ScanNet
\citep{dai2017scannet} scenes with 100 frames each: the first frame and
99 uniformly sampled remaining frames. On 7Scenes and NRGBD, we report
absolute trajectory error (ATE), translational and rotational
relative pose error (RPE-t/RPE-r), reconstruction accuracy (Acc), completeness
(Comp), and normal consistency (NC). Translation and reconstruction distances
are in meters; RPE-r is in degrees. Lower is better except for NC.
Results are sequence macro averages, displayed to three decimal places.
Boldface uses unrounded values.

\paragraph{Baselines and implementation.}
S55--S85 denote target sparsity of 55--85\% of patch-child interactions;
realized sparsity accounts for block rounding and always-kept support.
Within each backbone, baseline/VASC comparisons share checkpoints, inputs,
evaluators, geometry heads, and sparse kernels. We compare with
FasterVGGT~\citep{wang2026block} and our ${\pi}^3$ adaptation,
FasterPi3; tables label both FasterVGGT.
FastVGGT~\citep{shen2025fastvggt} uses its official reconstruction pipeline
at merge ratio 0.9. Timing covers full VGGT forward passes, including
geometry heads, at input size 518 and stride 10 on A6000 and RTX 4090.
We average per-sequence medians of three passes after one warm-up,
excluding loading and evaluation; comparisons use the same physical GPU.

\subsection{Main Results}

\input{tables/main_compact_strides}

\paragraph{Reconstruction and camera estimation.}
Table~\ref{tab:main} compares S65/S75 at strides 10 and 5 on both backbones.
VASC improves all three pose metrics over FasterVGGT and FasterPi3 in every
setting, and all six metrics at S75. On ${\pi}^3$, all six improve at
both sparsities and strides.

\begin{figure}[!htbp]
\centering
\includegraphics[width=\textwidth]{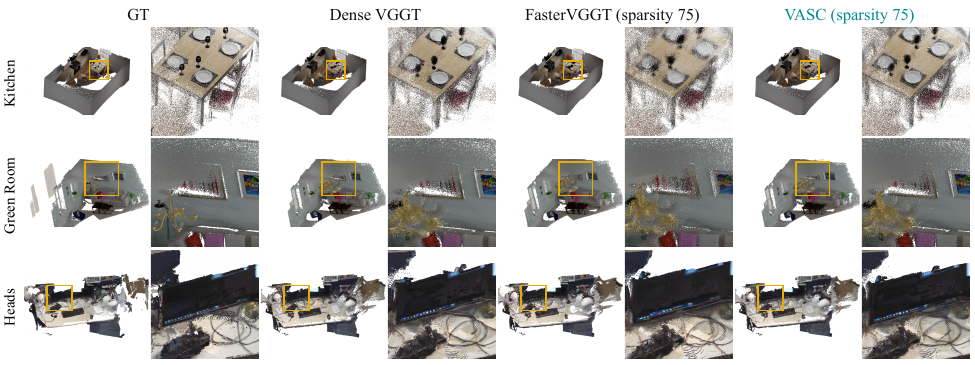}
\caption{\textbf{Qualitative reconstruction at sparsity 75.} Kitchen and Green Room
from NRGBD, and Heads from 7Scenes, using full stride-10 sequences.
Each method shows an overview and a detail crop; yellow boxes mark shared
regions. Cameras and crop bounds are shared after rigid alignment to GT.}
\label{fig:reconstruction}
\end{figure}

Figure~\ref{fig:reconstruction} compares local structure around tableware,
wall decorations, and monitor edges, complementing the broader
reconstruction comparisons in Figure~\ref{fig:teaser}.

\paragraph{Denser inputs and generalization.}
Gains persist with denser inputs: at stride 5 and S75 on NRGBD, VASC
improves VGGT ATE and Comp, and ${\pi}^3$ Acc and NC (Table~\ref{tab:main}).
On ScanNet-50, VASC at S75 reduces ATE, translational
RPE, and rotational RPE by 13.98\%, 14.40\%, and 18.10\%, respectively,
relative to FasterVGGT, while Chamfer distance remains nearly
unchanged.

\paragraph{Efficiency.}
On NRGBD at S85, VASC achieves $2.22\times$ and $2.29\times$ full-model
speedups over dense VGGT on A6000 and RTX 4090, respectively
(Table~\ref{tab:efficiency-main}), with 1.66--2.62\% latency overhead
over FasterVGGT.
\input{tables/efficiency_main}

\subsection{Ablation Study}

Table~\ref{tab:component-ablation} compares three variants at
75\% sparsity and frame stride 10: Value-aware only (no memory),
Faster+memory (value contrast set to one), and Full VASC.
Full VASC improves all six dataset-level metrics over
FasterVGGT on both backbones and datasets.

\begin{table}[!htb]
\centering
\footnotesize
\setlength{\tabcolsep}{2.5pt}
\caption{\textbf{S75 component ablation.}
Frame stride 10. FasterVGGT results are from Table~\ref{tab:main}.
Bold marks the best unrounded value within each backbone--dataset group.}
\label{tab:component-ablation}
\begin{tabular}{lllrrrrrr}
\toprule
Backbone & Dataset & Setting & ATE $\downarrow$ & RPE-t $\downarrow$ & RPE-r $\downarrow$ & Acc $\downarrow$ & Comp $\downarrow$ & NC $\uparrow$\\
\midrule
VGGT & 7Scenes & FasterVGGT & 0.084 & 0.038 & 1.080 & 0.018 & 0.028 & 0.646\\
& & Value-aware only & 0.078 & 0.034 & \textbf{0.957} & 0.018 & 0.028 & 0.649\\
& & Faster+memory & 0.078 & 0.035 & 0.975 & \textbf{0.018} & 0.028 & 0.648\\
& & Full VASC & \textbf{0.074} & \textbf{0.034} & 0.963 & 0.018 & \textbf{0.028} & \textbf{0.649}\\
\addlinespace
VGGT & NRGBD & FasterVGGT & 0.073 & 0.030 & 0.503 & 0.041 & 0.027 & 0.798\\
& & Value-aware only & \textbf{0.045} & 0.020 & 0.278 & 0.020 & 0.018 & \textbf{0.871}\\
& & Faster+memory & 0.049 & 0.022 & 0.325 & 0.023 & 0.020 & 0.848\\
& & Full VASC & 0.047 & \textbf{0.020} & \textbf{0.262} & \textbf{0.019} & \textbf{0.018} & 0.867\\
\midrule
Pi3 & 7Scenes & FasterVGGT & 0.066 & 0.027 & 0.855 & 0.031 & 0.027 & 0.709\\
& & Value-aware only & \textbf{0.061} & 0.026 & 0.815 & 0.025 & 0.025 & 0.743\\
& & Faster+memory & 0.064 & 0.026 & 0.827 & 0.025 & 0.026 & 0.753\\
& & Full VASC & 0.061 & \textbf{0.026} & \textbf{0.807} & \textbf{0.022} & \textbf{0.025} & \textbf{0.779}\\
\addlinespace
Pi3 & NRGBD & FasterVGGT & 0.074 & 0.020 & 0.277 & 0.056 & 0.033 & 0.779\\
& & Value-aware only & 0.062 & 0.018 & 0.228 & 0.042 & 0.029 & 0.831\\
& & Faster+memory & 0.064 & 0.019 & 0.226 & 0.040 & 0.028 & 0.849\\
& & Full VASC & \textbf{0.058} & \textbf{0.017} & \textbf{0.196} & \textbf{0.034} & \textbf{0.026} & \textbf{0.881}\\
\bottomrule
\end{tabular}
\end{table}

\begin{figure}[!htb]
\centering
\includegraphics[width=\textwidth]{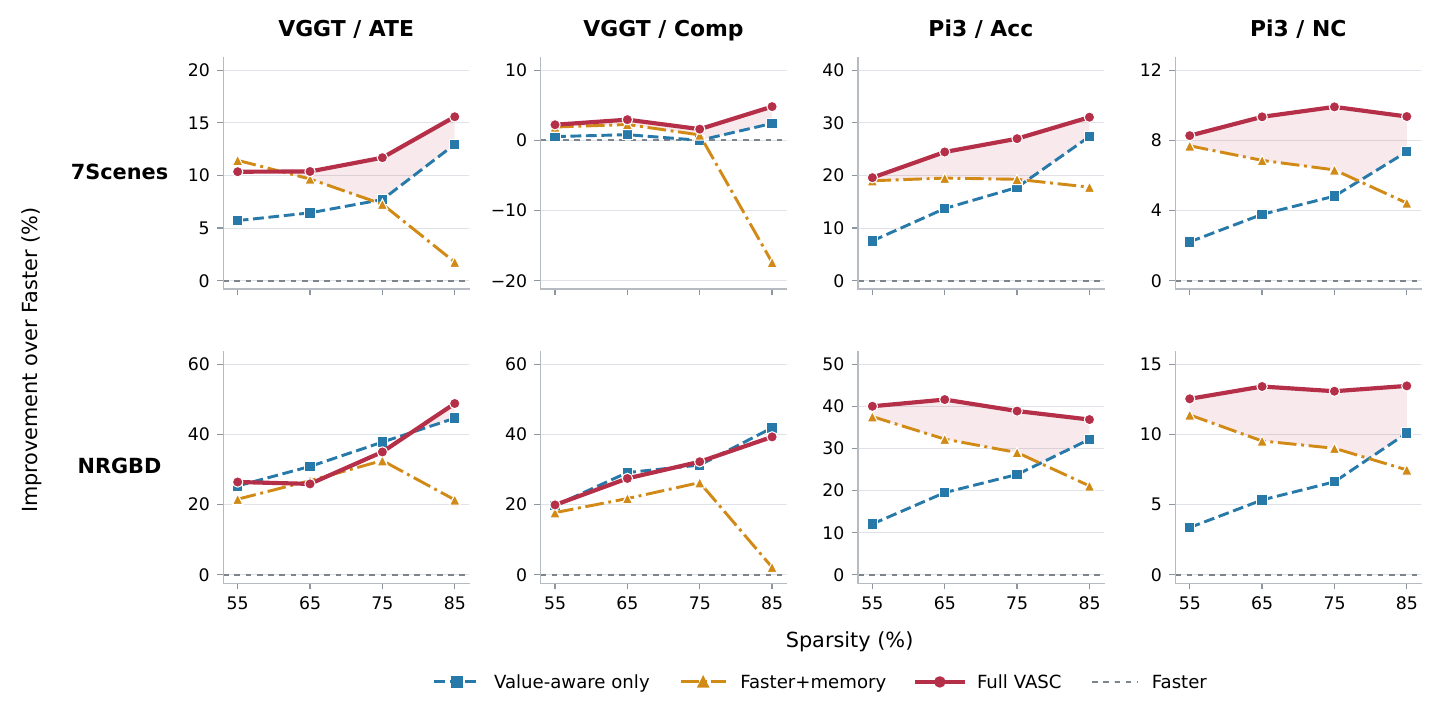}
\caption{\textbf{Component trends across sparsity.} Relative improvements
over FasterVGGT at stride 10; higher is better. Shading marks gains
over the stronger single component. Each panel uses its own y-axis scale.}
\label{fig:component-trends}
\end{figure}

\paragraph{Value-aware selection.}
The benefit of value-aware selection grows with sparsity on NRGBD.
Adding value contrast to Faster+memory reduces Acc by 5.20\%, 10.09\%,
15.32\%, and 49.07\% at S55, S65, S75, and S85 on VGGT; the corresponding
gains on ${\pi}^3$ are 3.96\%, 13.85\%, 13.89\%, and 19.97\%.
Value contrast delivers larger reconstruction gains on NRGBD as
the execution budget shrinks.

\paragraph{Cross-layer memory.}
At S75, adding memory to Value-aware only reduces ${\pi}^3$/NRGBD Acc
by 19.79\% and Comp by 11.08\%, while NC increases from 0.831 to 0.881.
Full VASC improves all six metrics over both single-component
variants on ${\pi}^3$/NRGBD. Relative to Value-aware only, memory improves
five of six metrics on ${\pi}^3$/7Scenes, with slightly higher ATE, and
five and four metrics on VGGT/7Scenes and VGGT/NRGBD, respectively,
with trade-offs in the remaining metrics.

\paragraph{Trends across sparsity levels.}
Figure~\ref{fig:component-trends} extends the component comparison to
S55--S85 on selected metrics. Improvements denote relative error
reductions for ATE, Acc, and Comp, and relative increases for NC.
Full VASC improves VGGT Comp over either
component at all four sparsities on 7Scenes and at S55/S75 on NRGBD;
on ${\pi}^3$, it improves Acc and NC across all four sparsities on both
datasets. The two components improve reconstruction together across
backbones and sparsity levels.

\paragraph{Memory adaptation to other schedulers.}
\label{sec:transfer}
We add memory to FastVGGT's token protection without changing its cosine
matcher, merge ratio (0.9), or token count, and to our AVGGT-R reproduction
at 85\% retained K/V service (Table~\ref{tab:host-memory-transfer}).
AVGGT-R uses the same selector with and without memory. Acc and Comp improve
on both datasets; FastVGGT Comp improves by 1.20\% while Acc worsens by
1.27\%. FastVGGT adaptation is evaluated within the 7Scenes collection
used for adapter selection.
\input{tables/host_transfer_summary}

%% file: tables/main_compact_strides.tex
\begin{table}[!htb]
\centering
\fontsize{8.5}{9}\selectfont
\setlength{\tabcolsep}{1pt}
\renewcommand{\arraystretch}{1.0}
\caption{\textbf{Comparison at strides 10 and 5.} Sequence means on 7Scenes
(18 sequences) and NRGBD (9). Bold marks the better unrounded value within each
FasterVGGT--VASC pair at the same backbone, stride, and sparsity.
FasterVGGT~\citep{wang2026block} is adapted to ${\pi}^3$ in its rows.
Dense is the unpruned reference; OOM: out of memory.}
\label{tab:main}
\begin{tabular}{@{}lccrrrrrr@{\hspace{5pt}}rrrrrr@{}}
\toprule[1.2pt]
Method & \shortstack{Sparsity\\(\%)} & Stride & \multicolumn{6}{c}{7Scenes} & \multicolumn{6}{c}{NeuralRGB-D}\\
\cmidrule(lr){4-9}\cmidrule(lr){10-15}
& & & ATE$\downarrow$ & RPE-t$\downarrow$ & RPE-r$\downarrow$ & Acc$\downarrow$ & Comp$\downarrow$ & NC$\uparrow$ & ATE$\downarrow$ & RPE-t$\downarrow$ & RPE-r$\downarrow$ & Acc$\downarrow$ & Comp$\downarrow$ & NC$\uparrow$\\
\midrule[0.6pt]
\multicolumn{15}{l}{\textit{VGGT}}\\
Dense & 0 & 10 & 0.069 & 0.031 & 0.859 & 0.017 & 0.028 & 0.652 & 0.030 & 0.016 & 0.179 & 0.014 & 0.015 & 0.908\\
FasterVGGT & 65 & 10 & 0.081 & 0.035 & 0.960 & \textbf{0.017} & 0.029 & 0.647 & 0.059 & 0.026 & 0.427 & 0.034 & 0.024 & 0.832\\
VASC & 65 & 10 & \textbf{0.073} & \textbf{0.033} & \textbf{0.911} & 0.018 & \textbf{0.028} & \textbf{0.650} & \textbf{0.044} & \textbf{0.019} & \textbf{0.239} & \textbf{0.018} & \textbf{0.017} & \textbf{0.880}\\
\cmidrule[0.3pt](l{6pt}r{6pt}){1-15}
FasterVGGT & 75 & 10 & 0.084 & 0.038 & 1.080 & 0.018 & 0.028 & 0.646 & 0.073 & 0.030 & 0.503 & 0.041 & 0.027 & 0.798\\
VASC & 75 & 10 & \textbf{0.074} & \textbf{0.034} & \textbf{0.963} & \textbf{0.018} & \textbf{0.028} & \textbf{0.649} & \textbf{0.047} & \textbf{0.020} & \textbf{0.262} & \textbf{0.019} & \textbf{0.018} & \textbf{0.867}\\
\midrule[0.75pt]
Dense & 0 & 5 & OOM & OOM & OOM & OOM & OOM & OOM & OOM & OOM & OOM & OOM & OOM & OOM\\
FasterVGGT & 65 & 5 & 0.081 & 0.024 & 0.682 & \textbf{0.016} & 0.027 & 0.610 & 0.066 & 0.024 & 0.383 & \textbf{0.030} & \textbf{0.021} & 0.817\\
VASC & 65 & 5 & \textbf{0.073} & \textbf{0.023} & \textbf{0.623} & 0.016 & \textbf{0.027} & \textbf{0.612} & \textbf{0.062} & \textbf{0.018} & \textbf{0.263} & 0.033 & 0.022 & \textbf{0.849}\\
\cmidrule[0.3pt](l{6pt}r{6pt}){1-15}
FasterVGGT & 75 & 5 & 0.084 & 0.027 & 0.757 & 0.017 & 0.027 & 0.608 & 0.076 & 0.027 & 0.446 & 0.031 & 0.022 & 0.787\\
VASC & 75 & 5 & \textbf{0.075} & \textbf{0.024} & \textbf{0.670} & \textbf{0.017} & \textbf{0.027} & \textbf{0.611} & \textbf{0.054} & \textbf{0.019} & \textbf{0.266} & \textbf{0.028} & \textbf{0.019} & \textbf{0.840}\\
\midrule[1.2pt]
\multicolumn{15}{l}{\textit{${\pi}^3$}}\\
Dense & 0 & 10 & 0.057 & 0.024 & 0.769 & 0.020 & 0.025 & 0.810 & 0.025 & 0.012 & 0.133 & 0.019 & 0.018 & 0.959\\
FasterVGGT & 65 & 10 & 0.064 & 0.026 & 0.835 & 0.029 & 0.026 & 0.719 & 0.063 & 0.019 & 0.252 & 0.050 & 0.031 & 0.794\\
VASC & 65 & 10 & \textbf{0.060} & \textbf{0.025} & \textbf{0.789} & \textbf{0.022} & \textbf{0.025} & \textbf{0.786} & \textbf{0.046} & \textbf{0.015} & \textbf{0.173} & \textbf{0.029} & \textbf{0.024} & \textbf{0.900}\\
\cmidrule[0.3pt](l{6pt}r{6pt}){1-15}
FasterVGGT & 75 & 10 & 0.066 & 0.027 & 0.855 & 0.031 & 0.027 & 0.709 & 0.074 & 0.020 & 0.277 & 0.056 & 0.033 & 0.779\\
VASC & 75 & 10 & \textbf{0.061} & \textbf{0.026} & \textbf{0.807} & \textbf{0.022} & \textbf{0.025} & \textbf{0.779} & \textbf{0.058} & \textbf{0.017} & \textbf{0.196} & \textbf{0.034} & \textbf{0.026} & \textbf{0.881}\\
\midrule[0.75pt]
Dense & 0 & 5 & 0.057 & 0.016 & 0.507 & 0.020 & 0.025 & 0.811 & 0.025 & 0.010 & 0.108 & 0.019 & 0.018 & 0.959\\
FasterVGGT & 65 & 5 & 0.064 & 0.017 & 0.535 & 0.029 & 0.026 & 0.719 & 0.060 & 0.015 & 0.196 & 0.050 & 0.030 & 0.793\\
VASC & 65 & 5 & \textbf{0.060} & \textbf{0.016} & \textbf{0.512} & \textbf{0.022} & \textbf{0.024} & \textbf{0.788} & \textbf{0.047} & \textbf{0.012} & \textbf{0.136} & \textbf{0.029} & \textbf{0.023} & \textbf{0.901}\\
\cmidrule[0.3pt](l{6pt}r{6pt}){1-15}
FasterVGGT & 75 & 5 & 0.066 & 0.018 & 0.544 & 0.030 & 0.026 & 0.709 & 0.069 & 0.016 & 0.213 & 0.055 & 0.032 & 0.779\\
VASC & 75 & 5 & \textbf{0.061} & \textbf{0.017} & \textbf{0.520} & \textbf{0.022} & \textbf{0.025} & \textbf{0.780} & \textbf{0.056} & \textbf{0.013} & \textbf{0.152} & \textbf{0.033} & \textbf{0.025} & \textbf{0.881}\\
\bottomrule[1.2pt]
\end{tabular}
\end{table}

%% file: tables/efficiency_main.tex
\begin{table}[!htbp]
\centering\scriptsize
\setlength{\tabcolsep}{4pt}
\caption{\textbf{Full-model efficiency on NeuralRGB-D.} Mean seconds per
complete stride-10 sequence. Speedup is relative to dense VGGT; overhead
is relative to FasterVGGT~\citep{wang2026block}.}
\label{tab:efficiency-main}
\begin{tabular}{lcrrrrr}
\toprule
GPU & Sparsity (\%) & Dense (s) & FasterVGGT (s) & VASC (s) & Speedup & Overhead\\
\midrule
A6000 & 75 & 23.849 & 11.981 & 12.239 & $1.950\times$ & 2.150\%\\
 & 85 & 23.849 & 10.506 & 10.739 & $2.220\times$ & 2.220\%\\
RTX 4090 & 75 & 15.631 & 7.256 & 7.446 & $2.100\times$ & 2.620\%\\
 & 85 & 15.631 & 6.650 & 6.822 & $2.290\times$ & 2.600\%\\
\bottomrule
\end{tabular}
\end{table}

%% file: tables/host_transfer_summary.tex
\begin{table}[!htbp]
\centering
\footnotesize
\setlength{\tabcolsep}{3pt}
\caption{\textbf{Memory adaptation across hosts.} All 18 7Scenes and nine
NRGBD sequences at stride 10; settings in Section~\ref{sec:transfer}.
FastVGGT follows \citet{shen2025fastvggt}.
Bold marks the better value in each pair.}
\label{tab:host-memory-transfer}
\begin{tabular}{lllrrrr}
\toprule
Host & Dataset & Configuration & Acc $\downarrow$ & Comp $\downarrow$ & NC1 $\uparrow$ & NC2 $\uparrow$\\
\midrule
FastVGGT & 7Scenes & Official & \textbf{0.018} & 0.027 & 0.616 & 0.648\\
& & Memory adapter & 0.018 & \textbf{0.027} & \textbf{0.617} & \textbf{0.650}\\
\midrule
AVGGT-R & 7Scenes & w/o memory & 0.018 & 0.028 & \textbf{0.647} & 0.650\\
& & w/ memory & \textbf{0.017} & \textbf{0.027} & 0.646 & \textbf{0.651}\\
\midrule
AVGGT-R & NRGBD & w/o memory & 0.020 & 0.018 & 0.844 & 0.841\\
& & w/ memory & \textbf{0.019} & \textbf{0.018} & \textbf{0.849} & \textbf{0.842}\\
\bottomrule
\end{tabular}
\end{table}

%% file: sections/05_conclusion.tex
\begin{samepage}
\section{Conclusion}

We presented VASC, a training-free method for efficient 3D reconstruction.
VASC combines pooled query--key relevance with adjacent-value contrast,
and uses cross-layer memory to carry unserved demand into subsequent
selection. Both components operate within a fixed budget while preserving
the native sparse kernel and geometry heads. Experiments on 7Scenes and
NeuralRGB-D with VGGT and ${\pi}^3$ demonstrate improved pose estimation
and reconstruction over sparse baselines, with up to $2.29\times$ faster
inference than dense VGGT.
\par
\end{samepage}

%% file: sections/appendix_a_implementation.tex
\section{Implementation Details}
\label{app:scheduler-details}

\subsection{Block Scores and Budgeted Selection}
At global layer $l$, $i$ indexes a pooled query block and $j$ a physical
key/value child block. Adjacent children in native patch-block order form
stable pairs $\mathcal C(p)=\{j_{p,1},j_{p,2}\}$; $p(j)$ identifies the
pair containing child $j$. A final unpaired child is a singleton. Pairing
is fixed across layers for a given layout. Batch and head indices are
omitted below.

Equations~\ref{eq:qk}--\ref{eq:pv} use average-pooled queries, keys,
and values with feature dimension $d$. RMS is the square root of the mean
squared feature entries, and
$\Pi_b$ is the proportional clipping operator defined below. The
neighboring-value cosine similarity in Section~\ref{sec:observation}
is a directional-variation diagnostic; it is not the RMS block score
used for selection.

For $N_q$ query blocks, demand and execution use identical averaging:
\begin{equation}
 \begin{aligned}
 a_l(p)&=\frac{1}{N_q|\mathcal C(p)|}
          \sum_i\sum_{j\in\mathcal C(p)} C_l(i,j),\\
 x_l(p)&=\frac{1}{N_q|\mathcal C(p)|}
          \sum_i\sum_{j\in\mathcal C(p)} X_l(i,j).
 \end{aligned}
 \label{eq:pair-demand}
\end{equation}
The realization of $\mathcal G$ broadcasts bounded pair memory across
queries and children, then selects within each query's fixed capacity:
\begin{equation}
 P_l(i,j)=C_l(i,j)+\widetilde M_l(p(j)).
 \label{eq:priority}
\end{equation}
\begin{equation}
 X_l(i,j)=\mathbf 1\!\left[j\in
 \operatorname{TopKIndices}_{b_l(i)}(P_l(i,:))\right],
 \qquad \sum_jX_l(i,j)=b_l(i).
 \label{eq:mask}
\end{equation}
The memory update uses the raw balance $M_l$, while selection uses only
the bounded copy $\widetilde M_l$.

\subsection{Proportional Clipping}
For a nonnegative vector $r$ and capacity $b\geq0$, the implementation uses
\begin{equation}
 [\Pi_b(r)]_j=\min\!\left(1,\frac{b r_j}{\sum_k r_k}\right),
 \qquad \Pi_b(0)=0.
 \label{eq:projection}
\end{equation}
This bounds each score by one and the total mass by $b$; saturation can
leave unused score mass. The binary Top-K mask in Equation~\ref{eq:mask}
separately enforces exact execution capacity. Memory uses
$c_l=\sum_p a_l(p)$ per batch and head, giving
$0\leq\widetilde M_l(p)\leq1$ and
$\sum_p\widetilde M_l(p)\leq c_l$.

\subsection{Value Contrast and Pairing}
For pooled values $u,v$ and scalar weights $\alpha,\beta$,
\begin{equation}
\alpha u+\beta v=(\alpha+\beta)\frac{u+v}{2}
 +(\alpha-\beta)\frac{u-v}{2}.
\label{eq:paired-value}
\end{equation}
The two terms separate common and differential content. RMS contrast
measures the latter, complementing query-dependent relevance; common
content can remain informative even at zero contrast. This interpretation
motivates selection rather than bounding attention-output error.
Pairs follow Section~\ref{sec:method}. An odd final child is paired with
itself for zero contrast and counted as a singleton in demand and service.

\subsection{Memory Normalization and Initialization}
For the settings studied here, the memory copy used for ranking is
\begin{equation}
 \widetilde M_l=\Pi_{c_l}(M_l),\qquad
 c_l=\sum_p a_l(p),\qquad M_1=\widetilde M_1=0.
 \label{eq:memory-normalization}
\end{equation}
Normalization is performed separately for each batch entry and head.
Only the ranking copy is normalized; the raw balance follows
Equation~\ref{eq:memory} and is carried to the next layer.

\subsection{Residual Accounting and State Lifetime}
Define excess service by $\rho_l(p)=[x_l(p)-M_l(p)-a_l(p)]_+$.
The nonnegative update in Equation~\ref{eq:memory} can then be written exactly as
\begin{equation}
 M_{l+1}=M_l+a_l-x_l+\rho_l.
\end{equation}
With $M_1=0$, summing gives
$M_{L+1}=\sum_{l=1}^{L}(a_l-x_l+\rho_l)$ pair-wise.
Since $a_l,x_l\in[0,1]$, induction gives
$0\leq M_l(p)\leq l-1$ for a fixed layout. The raw state retains accumulated
deficits; $\Pi_{c_l}$ bounds the copy used for ranking. Executing both
children supplies one unit per query before averaging, leaving a residual
if accumulated demand exceeds service.

\paragraph{Execution conventions.}
State is initialized per inference and reset on a layout mismatch.
A pending layer is settled upon entry to the next; the final layer is
settled after execution. The selected patch-child mask supplies $x_l$.
Camera and register queries retain native dense attention; special-only sink key
blocks retain the host's always-kept treatment. Handling of mixed boundary
blocks follows the native mask configuration and is shared with the
baseline. Selected child support is passed directly to native sparse
execution, with pose, depth, and point heads unchanged.

%% file: sections/appendix_b_observation_protocols.tex
\clearpage
\section{Observation and Visualization Protocols}
\label{app:observation-protocols}
We detail the intervention in Figure~\ref{fig:motivation-observation}
and the rendering protocol in Figure~\ref{fig:reconstruction}.

\input{sections/06_attention_only_observation_protocol}

\subsection{Qualitative Reconstruction Rendering}
Figure~\ref{fig:reconstruction} uses full stride-10 reconstructions of
Kitchen (152 frames), Green Room (145 frames), and Heads/seq-01
(100 frames); Figure~\ref{fig:teaser} uses Breakfast Room, Office/seq-06,
and Stairs/seq-04 (50 frames). Stairs uses S85; the other sparse
reconstructions and the teaser's dataset-level radar plots use S75.
Predictions are rigidly aligned to GT by point-to-point ICP (threshold 0.1,
at most 30 iterations, deterministic uniform registration samples).
All saved points are rendered with shared cameras, point footprints,
and crop bounds within each scene.

%% file: sections/06_attention_only_observation_protocol.tex
\subsection{Attention-Only Reconstruction and Local Value Analysis}
\label{app:attention-only-observation}

\paragraph{Reconstruction intervention.}
We compare dense and attention-only VGGT inference on
\texttt{office/seq-06} from 7Scenes (100 frames) and
\texttt{breakfast\_room} from NeuralRGB-D (117 frames), using full
stride-10 sequences. Each input image has resolution $518\times392$,
with a $28\times37$ patch grid and five special tokens. For each patch
query and head, the attention-only condition retains the highest-ranked
25\% of patch keys across all input views in each of the 24 global
layers. Attention is renormalized over retained patch and protected special
keys. Special queries, frame attention, weights, and prediction heads are
unchanged. Selection uses each layer's current hidden states, and the
resulting activations propagate through subsequent layers.

Controls include native dense inference and unmasked inference with
matched chunking. Figure~\ref{fig:motivation-observation} uses the latter,
with shared inputs, GT-defined camera, rendering settings, and tabletop
outline. This token-level diagnostic is separate from the FasterVGGT/VASC
benchmarks and is evaluated for reconstruction, not speed.

\paragraph{Attention and value maps.}
Figure~\ref{fig:motivation-observation}(b) uses Breakfast Room, global
layer 7 and head 10 (one-based), query patch $(14,18)$ in frame 58,
and key view 87 (zero-based coordinates). Pre-mask percentiles rank all
121,212 patch keys, including the query view but excluding special keys;
30,303 connections are retained per query and head. Higher percentiles
indicate stronger attention; red and gray show retained and omitted
connections. Value similarity averages cosine similarity to valid four-neighbors
in the same image and head. Display values are clipped to $[0.5,1]$;
statistics use unclipped values. Cosine describes directional variation,
whereas the method uses paired RMS contrast.

\paragraph{Spatial correspondence and selection.}
The 240-token crop spans rows $[6,22)$ and columns $[11,26)$.
Crop-relative boxes are A: $[0,5)\times[2,11)$,
B: $[6,10)\times[7,15)$, and C: $[10,16)\times[4,11)$.
Coordinates are shared across maps and mapped to the image at
$14\times14$ pixels per token. Boxes mark spatial regions;
Table~\ref{tab:observation-spatial-selection} reports their actual masks.

\begin{table}[H]
\centering
\small
\caption{\textbf{Regional attention, value similarity, and actual selection.}
Statistics for the spatial boxes in Figure~\ref{fig:motivation-observation}(b).
Retained counts refer to query--key connections in the illustrated head,
not tokens omitted from all network computation.}
\label{tab:observation-spatial-selection}
\begin{tabular}{lrrrr}
\toprule
Region & Keys & Retained & Mean attention percentile & Mean V cosine\\
\midrule
A & 45 & 9 & 54.500 & 0.610\\
B & 32 & 26 & 83.400 & 0.810\\
C & 42 & 8 & 43.900 & 0.660\\
\bottomrule
\end{tabular}
\end{table}

\paragraph{Selection and interpretation.}
The layer and head were chosen by inspecting saved maps; the three
regions were manually annotated. The maps illustrate an attention--value
mismatch, while reconstruction reflects the full-network intervention.
They are qualitative evidence, not a prevalence estimate or an isolated
head-level causal test. Matched-budget benchmarks evaluate method performance.

%% file: sections/appendix_c_benchmarks.tex
\clearpage
\section{Additional Benchmark Results}
\label{app:additional-experiments}
We report the full benchmark comparisons first, followed by denser input
sampling, ScanNet-50, the FastVGGT comparison, and forward timing.
Component controls are grouped separately in
Appendix~\ref{app:ablation-studies}.

\subsection{Complete Main Comparison}
\input{tables/main_complete}

\subsection{Denser Input Sampling}
\label{app:stride5}
The stride-5 evaluation retains every fifth frame, with no input cap.
\input{tables/stride5_complete}

\subsection{ScanNet-50}
\label{app:scannet50}
We evaluate the fixed 50-scene collection with 100 images per scene,
retaining the first frame and uniformly sampling the remaining frames
following the FastVGGT evaluation protocol. All three methods use identical
frame IDs within each scene, the same evaluator, depth-confidence threshold
1.0, and Chamfer distance cutoff 0.5. Unlike the stride-based benchmarks,
this protocol fixes the input count.
Both sparse methods use their S75 settings from Table~\ref{tab:main}.

\begin{table}[H]
\centering\small
\caption{\textbf{ScanNet-50 quality.} Macro means across all 50 scenes;
100 frames per scene. All metrics are lower-is-better.}
\label{tab:scannet50}
\begin{tabular}{lrrrrr}\toprule
Method & ATE & RPE-t & RPE-r & ARE & Chamfer\\\midrule
Dense VGGT & 0.092 & 0.055 & 1.206 & 3.950 & 0.441\\
FasterVGGT, S75 & 0.119 & 0.077 & 1.590 & 5.012 & 0.447\\
VASC, S75 & 0.102 & 0.066 & 1.303 & 4.253 & 0.447\\\midrule
VASC wins vs. FasterVGGT & 46/50 & 48/50 & 50/50 & 42/50 & 23/50\\
\bottomrule\end{tabular}
\end{table}

VASC improves all four pose means over FasterVGGT by 13.98\%, 14.40\%,
18.10\%, and 15.15\%, with scene-level wins reported in
Table~\ref{tab:scannet50}. Mean Chamfer is nearly unchanged (0.08\%
improvement), with improvements on 23 of 50 scenes.

\paragraph{Evaluation protocol.}
The two sparse methods are paired on the same host for every scene:
41 scenes on RTX 4090 and nine on RTX A6000.
For scene0056\_00, Dense runs on RTX 4090 and the sparse pair on RTX A6000;
input IDs and evaluator settings match.
These runs use one forward measurement without warm-up and report quality
only; latency follows the protocol in Appendix~\ref{app:timing}.

\clearpage
\subsection{Reconstruction Comparison with FastVGGT}
\label{app:fastvggt-comparison}
FastVGGT~\citep{shen2025fastvggt} uses the authors' implementation with merge ratio 0.9.
Its A6000 timings in Figure~\ref{fig:nc-latency} were measured separately
from the FasterVGGT/VASC timing runs.
Table~\ref{tab:fastvggt-medians} compares reconstruction medians at both
strides. On NRGBD, VASC improves Comp and NC medians over FastVGGT;
trade-offs remain in Acc and in 7Scenes NC at stride 5.
\input{tables/fastvggt_medians}

\subsection{Forward Timing}
\label{app:timing}
The timed operation is \texttt{model(images)}, including its geometry heads.
Loading, pose alignment, point-cloud evaluation, and ICP lie outside this region.
The evaluator's pose-only flag affects post-forward evaluation only;
all task heads remain in the timed forward.
Input size is 518 with native aspect-ratio handling and stride 10:
1,700 frames over 18 7Scenes sequences and 1,101 over nine NRGBD sequences.
One warm-up precedes three synchronized measured forwards per sequence.
We average sequence medians, and compute speedup as the ratio of dataset
mean Dense latency to dataset mean VASC latency. The baseline and VASC use
the same physical GPU, native kernel, and fixed-count selection.
A6000 has 48 GB and RTX 4090 has 24 GB of device memory.

\input{tables/timing_nopostcut}
\paragraph{Timing coverage.}
The table includes complete matched runs with no detected interference.
A6000 7Scenes VASC-S85 and RTX 4090 7Scenes Dense have no qualifying
completed records. The RTX 4090 7Scenes FasterVGGT-S75 interval is excluded
because of interference; none of these missing entries is imputed.

%% file: tables/main_complete.tex
\begin{table}[H]
\centering
\scriptsize
\setlength{\tabcolsep}{1.75pt}
\caption{\textbf{Complete stride-10 comparison on VGGT and ${\pi}^3$.} Dense is
the unpruned reference for each backbone. FasterVGGT~\citep{wang2026block}
denotes the original baseline on VGGT and our adaptation on ${\pi}^3$. Sparse baseline/VASC pairs have
matched realized capacity; bold marks the better raw value. Gain is VASC's
arithmetic mean over six direction-normalized relative changes, with metric
wins in parentheses. Every row uses complete native sequences (18/18 for
7Scenes, 9/9 for NRGBD).}
\label{tab:main-complete}
\resizebox{\textwidth}{!}{%
\begin{tabular}{llcrrrrrrr}
\toprule
Dataset & Method & S & ATE $\downarrow$ & RPE-t $\downarrow$ & RPE-r $\downarrow$ & Acc $\downarrow$ & Comp $\downarrow$ & NC $\uparrow$ & Gain\\
\midrule
7Scenes & Dense VGGT & 0 & 0.069 & 0.031 & 0.859 & 0.017 & 0.028 & 0.652 & --\\
& FasterVGGT & S55 & 0.079 & 0.033 & 0.922 & \textbf{0.017} & 0.028 & 0.650 & --\\
& VASC & S55 & \textbf{0.071} & \textbf{0.032} & \textbf{0.895} & 0.018 & \textbf{0.028} & \textbf{0.650} & $+2.500\%$ (5/6)\\
& FasterVGGT & S65 & 0.081 & 0.035 & 0.960 & \textbf{0.017} & 0.029 & 0.647 & --\\
& VASC & S65 & \textbf{0.073} & \textbf{0.033} & \textbf{0.911} & 0.018 & \textbf{0.028} & \textbf{0.650} & $+3.600\%$ (5/6)\\
& FasterVGGT & S75 & 0.084 & 0.038 & 1.080 & 0.018 & 0.028 & 0.646 & --\\
& VASC & S75 & \textbf{0.074} & \textbf{0.034} & \textbf{0.963} & \textbf{0.018} & \textbf{0.028} & \textbf{0.649} & $+6.200\%$ (6/6)\\
& FasterVGGT & S85 & 0.091 & 0.043 & 1.273 & 0.021 & 0.029 & 0.643 & --\\
& VASC & S85 & \textbf{0.077} & \textbf{0.037} & \textbf{1.028} & \textbf{0.019} & \textbf{0.028} & \textbf{0.646} & $+10.500\%$ (6/6)\\
\midrule
NRGBD & Dense VGGT & 0 & 0.030 & 0.016 & 0.179 & 0.014 & 0.015 & 0.908 & --\\
& FasterVGGT & S55 & 0.049 & 0.022 & 0.324 & 0.025 & 0.020 & 0.854 & --\\
& VASC & S55 & \textbf{0.036} & \textbf{0.018} & \textbf{0.203} & \textbf{0.016} & \textbf{0.016} & \textbf{0.888} & $+24.300\%$ (6/6)\\
& FasterVGGT & S65 & 0.059 & 0.026 & 0.427 & 0.034 & 0.024 & 0.832 & --\\
& VASC & S65 & \textbf{0.044} & \textbf{0.019} & \textbf{0.239} & \textbf{0.018} & \textbf{0.017} & \textbf{0.880} & $+29.200\%$ (6/6)\\
& FasterVGGT & S75 & 0.073 & 0.030 & 0.503 & 0.041 & 0.027 & 0.798 & --\\
& VASC & S75 & \textbf{0.047} & \textbf{0.020} & \textbf{0.262} & \textbf{0.019} & \textbf{0.018} & \textbf{0.867} & $+34.900\%$ (6/6)\\
& FasterVGGT & S85 & 0.118 & 0.040 & 0.650 & 0.057 & 0.036 & 0.757 & --\\
& VASC & S85 & \textbf{0.061} & \textbf{0.024} & \textbf{0.298} & \textbf{0.026} & \textbf{0.022} & \textbf{0.843} & $+41.300\%$ (6/6)\\
\midrule
7Scenes & Dense Pi3 & 0 & 0.057 & 0.024 & 0.769 & 0.020 & 0.025 & 0.810 & --\\
& FasterVGGT & S55 & 0.062 & 0.025 & 0.815 & 0.026 & 0.025 & 0.739 & --\\
& VASC & S55 & \textbf{0.058} & \textbf{0.024} & \textbf{0.774} & \textbf{0.021} & \textbf{0.024} & \textbf{0.800} & $+7.700\%$ (6/6)\\
& FasterVGGT & S65 & 0.064 & 0.026 & 0.835 & 0.029 & 0.026 & 0.719 & --\\
& VASC & S65 & \textbf{0.060} & \textbf{0.025} & \textbf{0.789} & \textbf{0.022} & \textbf{0.025} & \textbf{0.786} & $+9.400\%$ (6/6)\\
& FasterVGGT & S75 & 0.066 & 0.027 & 0.855 & 0.031 & 0.027 & 0.709 & --\\
& VASC & S75 & \textbf{0.061} & \textbf{0.026} & \textbf{0.807} & \textbf{0.022} & \textbf{0.025} & \textbf{0.779} & $+10.400\%$ (6/6)\\
& FasterVGGT & S85 & 0.071 & 0.029 & 0.902 & 0.034 & 0.028 & 0.692 & --\\
& VASC & S85 & \textbf{0.064} & \textbf{0.027} & \textbf{0.835} & \textbf{0.024} & \textbf{0.025} & \textbf{0.757} & $+12.300\%$ (6/6)\\
\midrule
NRGBD & Dense Pi3 & 0 & 0.025 & 0.012 & 0.133 & 0.019 & 0.018 & 0.959 & --\\
& FasterVGGT & S55 & 0.046 & 0.016 & 0.208 & 0.041 & 0.028 & 0.824 & --\\
& VASC & S55 & \textbf{0.037} & \textbf{0.014} & \textbf{0.154} & \textbf{0.025} & \textbf{0.021} & \textbf{0.927} & $+22.800\%$ (6/6)\\
& FasterVGGT & S65 & 0.063 & 0.019 & 0.252 & 0.050 & 0.031 & 0.794 & --\\
& VASC & S65 & \textbf{0.046} & \textbf{0.015} & \textbf{0.173} & \textbf{0.029} & \textbf{0.024} & \textbf{0.900} & $+25.600\%$ (6/6)\\
& FasterVGGT & S75 & 0.074 & 0.020 & 0.277 & 0.056 & 0.033 & 0.779 & --\\
& VASC & S75 & \textbf{0.058} & \textbf{0.017} & \textbf{0.196} & \textbf{0.034} & \textbf{0.026} & \textbf{0.881} & $+23.000\%$ (6/6)\\
& FasterVGGT & S85 & 0.107 & 0.023 & 0.344 & 0.068 & 0.040 & 0.751 & --\\
& VASC & S85 & \textbf{0.081} & \textbf{0.021} & \textbf{0.239} & \textbf{0.043} & \textbf{0.031} & \textbf{0.852} & $+23.100\%$ (6/6)\\
\bottomrule
\end{tabular}
}
\end{table}

%% file: tables/stride5_complete.tex
\begin{table}[H]
\centering
\footnotesize
\setlength{\tabcolsep}{2.0pt}
\caption{\textbf{Complete stride-5 comparison across sparsity.} FasterVGGT~\citep{wang2026block}
and VASC are paired at identical realized block capacity. Bold marks the better
raw value within each pair. Gain uses the same six-metric signed mean as
Table~\ref{tab:main-complete}, with wins in parentheses. All sequences are native and
uncapped (18/18 for 7Scenes, 9/9 for NRGBD).}
\label{tab:sparsity-kf}
\begin{tabular}{llcrrrrrrr}
\toprule
Dataset & Method & S & ATE $\downarrow$ & RPE-t $\downarrow$ & RPE-r $\downarrow$ & Acc $\downarrow$ & Comp $\downarrow$ & NC $\uparrow$ & Gain\\
\midrule
7Scenes & FasterVGGT & S55 & 0.080 & 0.024 & 0.650 & \textbf{0.016} & \textbf{0.027} & 0.611 & --\\
& VASC & S55 & \textbf{0.072} & \textbf{0.022} & \textbf{0.609} & 0.016 & 0.027 & \textbf{0.612} & $+3.000\%$ (4/6)\\
& FasterVGGT & S65 & 0.081 & 0.024 & 0.682 & \textbf{0.016} & 0.027 & 0.610 & --\\
& VASC & S65 & \textbf{0.073} & \textbf{0.023} & \textbf{0.623} & 0.016 & \textbf{0.027} & \textbf{0.612} & $+4.100\%$ (5/6)\\
& FasterVGGT & S75 & 0.084 & 0.027 & 0.757 & 0.017 & 0.027 & 0.608 & --\\
& VASC & S75 & \textbf{0.075} & \textbf{0.024} & \textbf{0.670} & \textbf{0.017} & \textbf{0.027} & \textbf{0.611} & $+6.200\%$ (6/6)\\
& FasterVGGT & S85 & 0.110 & 0.032 & 1.142 & 0.028 & 0.035 & 0.605 & --\\
& VASC & S85 & \textbf{0.077} & \textbf{0.027} & \textbf{0.737} & \textbf{0.018} & \textbf{0.027} & \textbf{0.608} & $+23.400\%$ (6/6)\\
\midrule
NRGBD & FasterVGGT & S55 & 0.056 & 0.022 & 0.339 & 0.028 & 0.019 & 0.835 & --\\
& VASC & S55 & \textbf{0.045} & \textbf{0.017} & \textbf{0.258} & \textbf{0.026} & \textbf{0.019} & \textbf{0.861} & $+11.800\%$ (6/6)\\
& FasterVGGT & S65 & 0.066 & 0.024 & 0.383 & \textbf{0.030} & \textbf{0.021} & 0.817 & --\\
& VASC & S65 & \textbf{0.062} & \textbf{0.018} & \textbf{0.263} & 0.033 & 0.022 & \textbf{0.849} & $+8.500\%$ (4/6)\\
& FasterVGGT & S75 & 0.076 & 0.027 & 0.446 & 0.031 & 0.022 & 0.787 & --\\
& VASC & S75 & \textbf{0.054} & \textbf{0.019} & \textbf{0.266} & \textbf{0.028} & \textbf{0.019} & \textbf{0.840} & $+21.700\%$ (6/6)\\
& FasterVGGT & S85 & 0.134 & 0.036 & 0.605 & 0.057 & 0.036 & 0.736 & --\\
& VASC & S85 & \textbf{0.067} & \textbf{0.020} & \textbf{0.268} & \textbf{0.031} & \textbf{0.023} & \textbf{0.820} & $+40.600\%$ (6/6)\\
\bottomrule
\end{tabular}
\end{table}

%% file: tables/fastvggt_medians.tex
\begin{table}[H]
\centering
\footnotesize
\setlength{\tabcolsep}{4pt}
\caption{\textbf{Reconstruction medians versus FastVGGT.} Sequence-averaged
point-level medians on all 18/9 sequences. FastVGGT~\citep{shen2025fastvggt} uses merge ratio 0.9;
VASC uses VGGT. Bold: best unrounded value per stride and dataset.}
\label{tab:fastvggt-medians}
\begin{tabular}{@{}lcrrr@{\hspace{10pt}}rrr@{}}
\toprule
Method & Stride & \multicolumn{3}{c}{7Scenes} & \multicolumn{3}{c}{NeuralRGB-D}\\
\cmidrule(lr){3-5}\cmidrule(lr){6-8}
& & Acc Med$\downarrow$ & Comp Med$\downarrow$ & NC Med$\uparrow$ & Acc Med$\downarrow$ & Comp Med$\downarrow$ & NC Med$\uparrow$\\
\midrule
FastVGGT & & 0.008 & 0.010 & 0.716 & \textbf{0.011} & 0.010 & 0.793\\
VASC (S65) & 10 & \textbf{0.007} & 0.010 & \textbf{0.733} & 0.011 & \textbf{0.007} & \textbf{0.974}\\
VASC (S75) & & 0.007 & \textbf{0.009} & 0.732 & 0.012 & 0.008 & 0.969\\
\midrule
FastVGGT & & 0.008 & 0.010 & \textbf{0.709} & \textbf{0.012} & 0.010 & 0.790\\
VASC (S65) & 5 & \textbf{0.006} & 0.008 & 0.674 & 0.018 & 0.008 & \textbf{0.945}\\
VASC (S75) & & 0.006 & \textbf{0.008} & 0.672 & 0.018 & \textbf{0.008} & 0.943\\
\bottomrule
\end{tabular}
\end{table}

%% file: tables/timing_nopostcut.tex
\begin{table}[H]
\centering\scriptsize
\setlength{\tabcolsep}{3pt}
\caption{\textbf{Full-model forward latency.} Seconds per complete stride-10 sequence, averaged over the dataset. Speedup is Dense/VASC; overhead is VASC/FasterVGGT$-1$~\citep{wang2026block}. Each sequence uses one warm-up and the median of three measured forwards. Only complete, uncontaminated matched comparisons are included.}
\label{tab:timing-current}
\resizebox{\linewidth}{!}{%
\begin{tabular}{llcrrrrr}
\toprule
GPU & Dataset & S & Dense (s) & FasterVGGT (s) & VASC (s) & Speedup & Overhead (\%)\\
\midrule
A6000 & 7Scenes & 55 & 14.865 & 9.951 & 10.124 & 1.470$\times$ & 1.740\\
A6000 & 7Scenes & 65 & 14.865 & 9.175 & 9.347 & 1.590$\times$ & 1.870\\
A6000 & 7Scenes & 75 & 14.865 & 8.373 & 8.528 & 1.740$\times$ & 1.850\\
A6000 & NRGBD & 55 & 23.849 & 14.830 & 15.076 & 1.580$\times$ & 1.660\\
A6000 & NRGBD & 65 & 23.849 & 13.402 & 13.666 & 1.750$\times$ & 1.970\\
A6000 & NRGBD & 75 & 23.849 & 11.981 & 12.239 & 1.950$\times$ & 2.150\\
A6000 & NRGBD & 85 & 23.849 & 10.506 & 10.739 & 2.220$\times$ & 2.220\\
RTX 4090 & NRGBD & 55 & 15.631 & 8.468 & 8.675 & 1.800$\times$ & 2.450\\
RTX 4090 & NRGBD & 65 & 15.631 & 7.852 & 8.052 & 1.940$\times$ & 2.560\\
RTX 4090 & NRGBD & 75 & 15.631 & 7.256 & 7.446 & 2.100$\times$ & 2.620\\
RTX 4090 & NRGBD & 85 & 15.631 & 6.650 & 6.822 & 2.290$\times$ & 2.600\\
\bottomrule
\end{tabular}}
\end{table}

%% file: sections/appendix_d_ablations.tex
\clearpage
\section{Component Ablations}
\label{app:ablation-studies}
The component study examines value-aware scoring and cross-layer memory.
The three component variants share the same implementation and all
settings except the component being varied. FasterVGGT reference rows in
Table~\ref{tab:component-ablation} are taken from the main benchmark runs
in Table~\ref{tab:main}; they are not an additional both-components-off
run of the ablation implementation.

\label{app:component-matrix}
Tables~\ref{tab:components-vggt-7scenes}--\ref{tab:components-pi3-nrgbd}
report all six metrics at stride 10, averaged over all 18 7Scenes or nine
NeuralRGB-D sequences. Value-aware only removes memory; Faster+memory
sets paired-value contrast to one. S denotes sparsity (\%). Bold marks
the best unrounded value in each three-row group, including when displayed
values coincide after rounding.

\subsection{VGGT}

\begin{table}[H]\centering\small
\caption{\textbf{VGGT components on 7Scenes.} Frame stride 10.}
\label{tab:components-vggt-7scenes}
\begin{tabular*}{\textwidth}{@{\extracolsep{\fill}}clrrrrrr@{}}\toprule
S & Variant & ATE $\downarrow$ & RPE-t $\downarrow$ & RPE-r $\downarrow$ & Acc $\downarrow$ & Comp $\downarrow$ & NC $\uparrow$\\\midrule
55 & Value-aware only & 0.075 & 0.032 & 0.886 & 0.017 & 0.028 & 0.650\\
 & Faster+memory & \textbf{0.070} & \textbf{0.031} & \textbf{0.873} & \textbf{0.017} & 0.028 & 0.650\\
 & Full VASC & 0.071 & 0.032 & 0.895 & 0.018 & \textbf{0.028} & \textbf{0.650}\\
\addlinespace
65 & Value-aware only & 0.076 & \textbf{0.033} & \textbf{0.910} & \textbf{0.017} & 0.028 & 0.649\\
 & Faster+memory & 0.074 & 0.033 & 0.927 & 0.018 & 0.028 & 0.649\\
 & Full VASC & \textbf{0.073} & 0.033 & 0.911 & 0.018 & \textbf{0.028} & \textbf{0.650}\\
\addlinespace
75 & Value-aware only & 0.078 & 0.034 & \textbf{0.957} & 0.018 & 0.028 & 0.649\\
 & Faster+memory & 0.078 & 0.035 & 0.975 & \textbf{0.018} & 0.028 & 0.648\\
 & Full VASC & \textbf{0.074} & \textbf{0.034} & 0.963 & 0.018 & \textbf{0.028} & \textbf{0.649}\\
\addlinespace
85 & Value-aware only & 0.079 & \textbf{0.037} & 1.030 & \textbf{0.019} & 0.029 & \textbf{0.646}\\
 & Faster+memory & 0.090 & 0.042 & 1.232 & 0.023 & 0.035 & 0.644\\
 & Full VASC & \textbf{0.077} & 0.037 & \textbf{1.028} & 0.019 & \textbf{0.028} & 0.646\\
\bottomrule\end{tabular*}\end{table}
\begin{table}[H]\centering\small
\caption{\textbf{VGGT components on NeuralRGB-D.} Frame stride 10.}
\label{tab:components-vggt-nrgbd}
\begin{tabular*}{\textwidth}{@{\extracolsep{\fill}}clrrrrrr@{}}\toprule
S & Variant & ATE $\downarrow$ & RPE-t $\downarrow$ & RPE-r $\downarrow$ & Acc $\downarrow$ & Comp $\downarrow$ & NC $\uparrow$\\\midrule
55 & Value-aware only & 0.036 & 0.018 & 0.228 & 0.017 & 0.016 & \textbf{0.888}\\
 & Faster+memory & 0.038 & 0.018 & 0.212 & 0.016 & 0.016 & 0.884\\
 & Full VASC & \textbf{0.036} & \textbf{0.018} & \textbf{0.203} & \textbf{0.016} & \textbf{0.016} & 0.888\\
\addlinespace
65 & Value-aware only & \textbf{0.041} & 0.019 & 0.254 & 0.018 & \textbf{0.017} & \textbf{0.880}\\
 & Faster+memory & 0.043 & 0.020 & 0.253 & 0.020 & 0.019 & 0.871\\
 & Full VASC & 0.044 & \textbf{0.019} & \textbf{0.239} & \textbf{0.018} & 0.017 & 0.880\\
\addlinespace
75 & Value-aware only & \textbf{0.045} & 0.020 & 0.278 & 0.020 & 0.018 & \textbf{0.871}\\
 & Faster+memory & 0.049 & 0.022 & 0.325 & 0.023 & 0.020 & 0.848\\
 & Full VASC & 0.047 & \textbf{0.020} & \textbf{0.262} & \textbf{0.019} & \textbf{0.018} & 0.867\\
\addlinespace
85 & Value-aware only & 0.066 & \textbf{0.023} & 0.325 & \textbf{0.024} & \textbf{0.021} & \textbf{0.845}\\
 & Faster+memory & 0.093 & 0.031 & 0.456 & 0.051 & 0.036 & 0.814\\
 & Full VASC & \textbf{0.061} & 0.024 & \textbf{0.298} & 0.026 & 0.022 & 0.843\\
\bottomrule\end{tabular*}\end{table}
\clearpage
\subsection{\texorpdfstring{${\pi}^3$}{Pi3}}
\begin{table}[H]\centering\small
\caption{\textbf{${\pi}^3$ components on 7Scenes.} Frame stride 10.}
\label{tab:components-pi3-7scenes}
\begin{tabular*}{\textwidth}{@{\extracolsep{\fill}}clrrrrrr@{}}\toprule
S & Variant & ATE $\downarrow$ & RPE-t $\downarrow$ & RPE-r $\downarrow$ & Acc $\downarrow$ & Comp $\downarrow$ & NC $\uparrow$\\\midrule
55 & Value-aware only & 0.060 & 0.025 & 0.801 & 0.024 & 0.025 & 0.755\\
 & Faster+memory & 0.059 & 0.025 & \textbf{0.774} & 0.021 & \textbf{0.024} & 0.795\\
 & Full VASC & \textbf{0.058} & \textbf{0.024} & 0.774 & \textbf{0.021} & 0.024 & \textbf{0.800}\\
\addlinespace
65 & Value-aware only & 0.061 & 0.025 & 0.811 & 0.025 & 0.025 & 0.746\\
 & Faster+memory & 0.063 & 0.026 & 0.803 & 0.023 & 0.025 & 0.768\\
 & Full VASC & \textbf{0.060} & \textbf{0.025} & \textbf{0.789} & \textbf{0.022} & \textbf{0.025} & \textbf{0.786}\\
\addlinespace
75 & Value-aware only & \textbf{0.061} & 0.026 & 0.815 & 0.025 & 0.025 & 0.743\\
 & Faster+memory & 0.064 & 0.026 & 0.827 & 0.025 & 0.026 & 0.753\\
 & Full VASC & 0.061 & \textbf{0.026} & \textbf{0.807} & \textbf{0.022} & \textbf{0.025} & \textbf{0.779}\\
\addlinespace
85 & Value-aware only & \textbf{0.062} & \textbf{0.027} & \textbf{0.833} & 0.025 & 0.025 & 0.743\\
 & Faster+memory & 0.068 & 0.029 & 0.904 & 0.028 & 0.027 & 0.723\\
 & Full VASC & 0.064 & 0.027 & 0.835 & \textbf{0.024} & \textbf{0.025} & \textbf{0.757}\\
\bottomrule\end{tabular*}\end{table}
\begin{table}[H]\centering\small
\caption{\textbf{${\pi}^3$ components on NeuralRGB-D.} Frame stride 10.}
\label{tab:components-pi3-nrgbd}
\begin{tabular*}{\textwidth}{@{\extracolsep{\fill}}clrrrrrr@{}}\toprule
S & Variant & ATE $\downarrow$ & RPE-t $\downarrow$ & RPE-r $\downarrow$ & Acc $\downarrow$ & Comp $\downarrow$ & NC $\uparrow$\\\midrule
55 & Value-aware only & 0.042 & 0.016 & 0.194 & 0.036 & 0.026 & 0.852\\
 & Faster+memory & 0.042 & 0.015 & 0.160 & 0.026 & 0.022 & 0.918\\
 & Full VASC & \textbf{0.037} & \textbf{0.014} & \textbf{0.154} & \textbf{0.025} & \textbf{0.021} & \textbf{0.927}\\
\addlinespace
65 & Value-aware only & 0.054 & 0.017 & 0.217 & 0.040 & 0.028 & 0.836\\
 & Faster+memory & 0.056 & 0.017 & 0.196 & 0.034 & 0.025 & 0.869\\
 & Full VASC & \textbf{0.046} & \textbf{0.015} & \textbf{0.173} & \textbf{0.029} & \textbf{0.024} & \textbf{0.900}\\
\addlinespace
75 & Value-aware only & 0.062 & 0.018 & 0.228 & 0.042 & 0.029 & 0.831\\
 & Faster+memory & 0.064 & 0.019 & 0.226 & 0.040 & 0.028 & 0.849\\
 & Full VASC & \textbf{0.058} & \textbf{0.017} & \textbf{0.196} & \textbf{0.034} & \textbf{0.026} & \textbf{0.881}\\
\addlinespace
85 & Value-aware only & 0.081 & \textbf{0.020} & 0.250 & 0.046 & 0.032 & 0.827\\
 & Faster+memory & 0.089 & 0.022 & 0.291 & 0.053 & 0.036 & 0.807\\
 & Full VASC & \textbf{0.081} & 0.021 & \textbf{0.239} & \textbf{0.043} & \textbf{0.031} & \textbf{0.852}\\
\bottomrule\end{tabular*}\end{table}

%% file: sections/appendix_e_adapters.tex
\clearpage
\section{Memory Adaptation to Other Attention-Reduction Methods}
\label{app:host-adapters}
Table~\ref{tab:host-memory-transfer} summarizes memory adaptation to
FastVGGT and AVGGT-R, retaining each host's attention-reduction mechanism.
\subsection{FastVGGT}
The adapter derives arrivals from the native source-to-destination merge
residual and counts independent source retention as service. It modifies
source-protection priority while preserving destination-protection slots,
the cosine matcher, merge ratio 0.9, and realized attention-token count.
We use current scale 5, memory weight 0.25, and arrival load 1.0,
selected on two four-sequence subsets of the evaluated 18-sequence
7Scenes collection. Results characterize in-collection adaptation,
rather than held-out generalization.

\begin{table}[H]\centering\small
\caption{\textbf{FastVGGT adaptation on 7Scenes.} Frame stride 10.
Means are macro averages of per-sequence statistics; ``median'' rows
first take the within-sequence median, then average across sequences.}
\label{tab:fastvggt-adapter-details}
\begin{tabular}{lrrrr}\toprule
Variant & Acc $\downarrow$ & Comp $\downarrow$ & NC1 $\uparrow$ & NC2 $\uparrow$\\\midrule
Official, mean & 0.018 & 0.027 & 0.616 & 0.648\\
Adapted, mean & 0.018 & 0.027 & 0.617 & 0.650\\
Official, median & 0.008 & 0.010 & 0.682 & 0.750\\
Adapted, median & 0.008 & 0.010 & 0.684 & 0.754\\
\bottomrule\end{tabular}\end{table}

\subsection{AVGGT-R}
AVGGT-R is our independent reproduction. Memory starts at the first
retained global layer (zero-based layer 9) and
propagates through the remaining 15 global layers. Both arms use the same
QK-derived current priority, factor-4 retained K/V grid, first-frame
retention, diagonal correction, and omitted-token mean correction.
The 85\% service-retention ratio applies to this host's retained service
support; it is distinct from the S85 patch sparsity in the main experiments.
No paired-value factor is added.

\paragraph{Reconstruction and pose.}
On 7Scenes, Acc and Comp improve by 3.11\%
and 1.72\%; on NeuralRGB-D they improve by 6.44\% and 3.80\%.
At 90\% retained service on 7Scenes, all eight reconstruction aggregates
improve, while the three pose means regress by 0.09--0.43\%.